\documentclass[10pt,twocolumn,letterpaper]{article}

\usepackage[pagenumbers]{cvpr} % To force page numbers, e.g. for an arXiv version

\usepackage{graphicx}
\usepackage{amsmath}
\usepackage{amssymb}
\usepackage{booktabs}
\usepackage[pagebackref,breaklinks,colorlinks]{hyperref}

\usepackage[capitalize]{cleveref}
\crefname{section}{Sec.}{Secs.}
\Crefname{section}{Section}{Sections}
\Crefname{table}{Table}{Tables}
\crefname{table}{Tab.}{Tabs.}

\def\cvprPaperID{*****} % *** Enter the CVPR Paper ID here
\def\confName{CVPR}
\def\confYear{2022}

\begin{document}

% \title{\LaTeX\ Author Guidelines for \confName~Proceedings}
\title{Learning Regularized Multi-Scale Feature Flow for\\ High Dynamic Range Imaging} % Replace with your title

\author{
    % Authors
    Qian Ye\textsuperscript{\rm 1},
    Masanori Suganuma\textsuperscript{\rm 1,2},
    Jun Xiao\textsuperscript{\rm 3},
    Takayuki Okatani\textsuperscript{\rm 1,2}\\
    \textsuperscript{\rm 1} Graduate School of Information Sciences, Tohoku University \\
    \textsuperscript{\rm 2} RIKEN Center for AIP \\
    \textsuperscript{\rm 3} Department of Electronic and Information Engineering, The Hong Kong Polytechnic University \\
    \{qian, suganuma, okatani\}@vision.is.tohoku.ac.jp ~ jun.xiao@connect.polyu.hk
}

% \author{
%     % Affiliations
%     \textsuperscript{\rm 1} Graduate School of Information Sciences, Tohoku University \\
%     \textsuperscript{\rm 2} RIKEN Center for AIP \\
%     \{zhijie, ryu, suganuma, okatani\}@vision.is.tohoku.ac.jp
% }
% \author{Qian Ye\\
% % Institution1\\
% Graduate School of Information Sciences,\\ Tohoku University\\
% {\tt\small qian@vision.is.tohoku.ac.jp}
% \and
% Jun Xiao\\
% Department of Electronic and Information Engineering,\\ The Hong Kong Polytechnic University\\
% {\tt\small jun.xiao@connect.polyu.hk}
% }
\maketitle

%%%%%%%%% ABSTRACT
\begin{abstract}
Reconstructing ghosting-free high dynamic range (HDR) images of dynamic scenes from a set of multi-exposure images is a challenging task, especially with large object motion and occlusions, leading to visible artifacts using existing methods. To address this problem, we propose a deep network that tries to learn multi-scale feature flow guided by the regularized loss. It first extracts multi-scale features and then aligns features from non-reference images. After alignment, we use residual channel attention blocks to merge the features from different images. Extensive qualitative and quantitative comparisons show that our approach achieves state-of-the-art performance and produces excellent results where color artifacts and geometric distortions are significantly reduced.
\end{abstract}

%%%%%%%%% BODY TEXT
\section{Introduction}
\label{sec:intro}

High dynamic range (HDR) imaging is a method to generate a larger dynamic range of illumination than standard imaging systems. It has been applied to movies \cite{rempel2009video} and computer rendering \cite{anderson2007critters} to gain more information and better visual experience. As cameras that can capture HDR images are generally expensive, an alternative way to get HDR images is to reconstruct HDR images from a series of low dynamic range (LDR) images captured by a standard camera with different exposure settings. While they can reconstruct high-quality HDR images for static scenes, the existing methods tend to yield images with many ghosting artifacts for dynamic scenes, in which the imaging scenes are static captured by a hand-held camera or there are some moving objects.
% Recently, learning-based methods using convolutional neural networks (CNNs) have been shown to generate sufficient HDR images. 

% Various methods have been developed to address the ghosting issues, such as those based on detecting and rejecting moving regions \cite{khan2006ghost,heo2010ghost}, optical flow \cite{bogoni2000extending,hu2013hdr,hafner2014simultaneous}, patch-based registration \cite{sen2012robust,ma2017robust}, etc. These methods improve the robustness against the ghosting artifacts, but they often fail for the scenes with large motions and brightness changes.
Increasing efforts have been invested in exploring how to remove ghosting artifacts in the multi-exposure-based HDR reconstruction. There are several methods that attempt to detect motion regions in the input LDR images and then remove these regions in the step of merging the images \cite{khan2006ghost,heo2010ghost,yan2017high}. However, they tend to work well only when the motion in the input images is relatively small. When there are large motions, a large number of image pixels need to be removed, which results in incorrect reconstruction because the information about these pixels is lost. 

% Recently, learning-based methods employing convolutional neural networks (CNNs) have been introduced to overcome the limitation of the previous work leading to remarkable results. 
%In the typical HDR imaging strategy, we select one of the LDR images as a reference image, and methods generate an HDR image corresponding to the selected reference image.
% In the typical HDR imaging strategy, an LDR image is selected from a series of the captured LDR images as the reference, and then the corresponding HDR image is generated. The key to success lies in the problem of how we align and merge non-reference LDR images and a reference LDR image. Researchers have considered several approaches to the problem. Kalantari \etal first align the LDR images using optical flow and then apply CNNs to merge the LDR images \cite{kalantari2017deep}. 

% Yan \etal propose an attention-guided network to learn to identify misaligned elements before merging the LDR images \cite{yan2019attention}. Pu \etal introduce a deformable convolution to suppress misaligned features and fully utilize feature information \cite{pu2020robust}. Although they have achieved better performance than the previous methods, they still suffer from the ghosting artifacts when significant motion is in the input LDR images (see examples in Fig.\ref{fig:ghosting}). 
\begin{figure}[t]
\begin{center}
\includegraphics[width=0.95\linewidth]{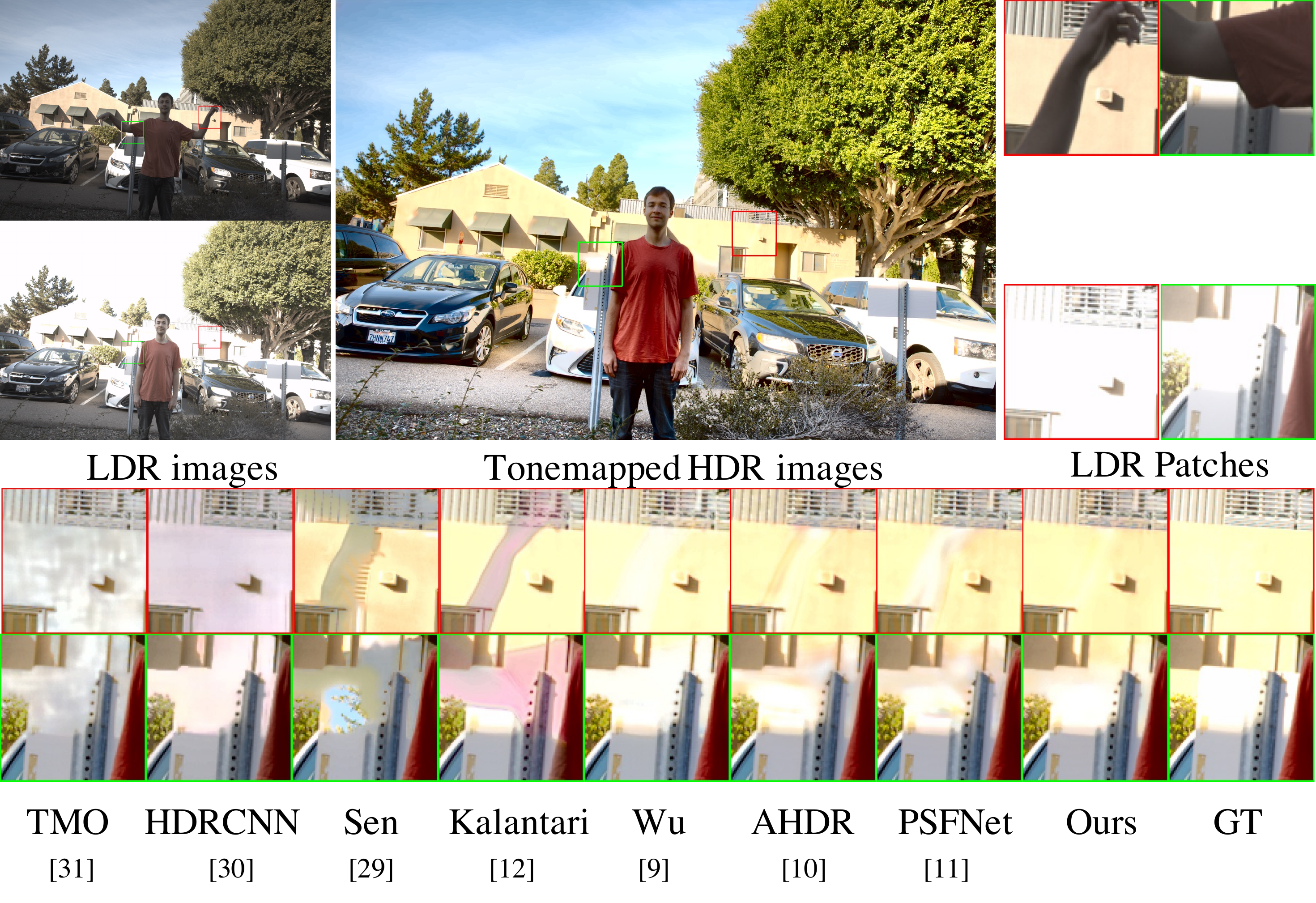}
\end{center}
\vspace*{-4mm}
\caption{Examples of generated HDR images from the test set of \cite{kalantari2017deep}. The zoomed regions of different methods are highlighted.}

%\vspace*{-8mm}
\label{fig:ghosting}
\end{figure}

Another approach has also been studied, which is to align input LDR images to a reference view, and then merge them altogether for HDR image reconstruction \cite{bogoni2000extending,hu2013hdr,hafner2014simultaneous}. Many recent methods employ convolutional neural network (CNN) to improve reconstruction image quality. However, there is still room for improvements in the area of ghosting artifacts. End-to-end learning-based approaches such as \cite{wu2018deep,yan2019attention,ye2021progressive} without implicit alignment directly feed LDR images into a network to reconstruct HDR images, failing to deal with scenarios with complex motion or large disparity. As shown in Fig. \ref{fig:ghosting}, these end-to-end leaning-based methods fails to deal with the motion region. 
The method in \cite{kalantari2017deep} performs optical flow-based image alignment followed by a convolutional neural network at the merging process. Aligning images in the pixel domain is often prone to noisy or saturated pixels-induced misalignment, which leads to visible artifacts in the final synthesized presentation. In addition, the classical optical flow methods and the optical flow models pre-trained on other datasets can not deal with the occlusion region in which the ghosting artifacts often occur. As suggested in \cite{chan2021basicvsr}, feature warping can achieve better performance compared with warping the image. The method in \cite{pu2020robust} performs alignment in feature domain by using deformable convolution layers \cite{zhu2019deformable}. However, it has a limitation in finding long-distance correspondence; as argued in \cite{lin2021fdan}, deformable convolution could also lead to an unstable training process and limited generalization. Inspired by non-local structure \cite{wang2018non}, Choi \etal \cite{choi2020pyramid} proposed to calculate the inter-similarity between LDR images for every pixel, which are used to align non-reference features toward the reference feature. However, this non-local structure-based operation is computationally expensive and needs large memory when the size of input images is large, while the images for HDR imaging often have a large size.
According to \cite{xue2019video}, Task-Oriented flow learns to handle occlusions well, though its estimated
motion field differs from the ground truth optical flow. Considered the benefit from Task-Orient flow, Kalantari \etal \cite{kalantari2019deep} proposed a Task-Oriented flow network which is specifically designed for HDR video reconstruction and is only based on the loss for HDR video reconstruction. This Task-Oriented flow network performs better than pre-trained or classical optimization-based optical flow methods since it can deal with the occlusion, which reduces the artifacts in occlusion regions. However, only trained on the task-specific loss (\ie HDR reconstruction loss), the Task-Oriented flow will fail on the large saturated areas in which there are a few details. In this case, this misalignment leads to artifacts on the over-exposed regions in the reference image.

Inspired by the photometric loss \cite{yu2016back} for self-supervised learning of optical flow, we proposed the regularized loss to provide supervision for flow learning to address the misalignment in Task-Orient flow in HDR imaging. We directly reconstruct an HDR image based on the aligned features and compute loss between this reconstructed image and the corresponding ground truth.  

Differing from the previous methods \cite{kalantari2017deep,kalantari2019deep,prabhakar2019fast} that use the existing optical flow models like SPyNet~\cite{ranjan2017optical} and PWC-Net~\cite{sun2018pwc}, we design a simple but effective network for learning the flow for feature alignment. We remove the context encoder in the flow network and directly use the features for HDR image reconstruction as the input for flow estimation. We argue that this flow structure can achieve better alignment performance in the HDR imaging task since there are large illumination changes and large saturated areas in image space, while the extracted features for HDR reconstruction can provide rich information to avoid misalignment. And we name this flow structure as feature flow. 

To this end, we propose a new network that enables end-to-end training, including alignment. The proposed method consists of two networks: {\em alignment network} and {\em merging network} in this order. The alignment network extracts multi-scale features from the input LDR images and estimates optical flow. It then aligns the non-reference LDR images to the reference LDR image in feature space using the estimated flow. 
The merging network takes the aligned features and multi-scale features as input and generates a final HDR image using a residual attention mechanism. Experimental results show that our method can achieve better quantitative and qualitative evaluation performance than the existing state-of-the-art methods on the commonly used public test datasets. 

%-------------------------------------------------------------------------
%------------------------------------------------------------------------- 
\section{Related Works}
\subsection{Motion Removal based Methods}
% {\color{red}Motion detection and motion removal are key points in motion-removal-based methods. Specifically,} 
These methods is firstly to detect the motion region and then remove these pixels on the motion region in the merging processing. 
% These methods classify each pixel in the images as belonging to a static region or moving object. They then only merge the pixels classified as static while rejecting those classified as moving.
Khan \etal \cite{khan2006ghost} use a non-parametric model to compute weights iteratively and apply these computed weights to pixels to fuse multiple LDR images to obtain final HDR images. Heo \etal \cite{heo2010ghost} detect motion regions based on the joint probability densities and refine these regions by using energy minimization. Jacobs \etal \cite{jacobs2008automatic} propose a method to detect moving pixels based on the difference between the LDR images. Zhang \etal \cite{zhang2011gradient} propose motion detection method based on the image gradients between different images. 
% Lee \etal \cite{lee2014ghost} consider the motion detection problem as a rank-minimization problem while Yan \etal \cite{yan2017high} use sparse representation to detect the motion. 
Lee \etal \cite{lee2014ghost} considered that the noise, moving objects, and distortions as outliers, so they proposed a low-rank model to reconstruct HDR images. Following their method, Yan \etal \cite{yan2017high} proposed a sparse model to detect motion regions. When the motion in LDR images is small, motion-removal-based methods can achieve satisfactory results. However, when the motion is large, a large number of pixels are unavoidably removed in the merging stage, causing undesirable artifacts in the generated HDR images.

\subsection{Alignment based Methods}
Most alignment based methods adopt optical flow and its variants to align LDR images and then merge aligned LDR images to generate corresponding HDR images. 
% These methods align input LDR images and then merge the aligned LDR images to generate an HDR image. 
% Most of these methods adopt optical flow and its variants for alignment.
% Bogoni \cite{bogoni2000extending} estimates motion vectors using optical flow and uses the estimated field for the alignment by warping the input images with them. 
Bogoni \cite{bogoni2000extending} use optical flow to estimate motion filed between LDR images and then warped and aligned these LDR images by using the computed motion field.
% In \cite{kang2003high}, intensities of LDR images are transformed into the luminance domain using exposure time information and a method for merging images is proposed to eliminate the artifacts of the alignment using optical flow. 
Instead of fusing LDR images in the spatial domain, Kang \etal \cite{kang2003high} firstly utilize the information of exposure time and converted LDR images into luminance domain. In the fusion process, a method was proposed to eliminate artifacts by using the optical flow.
Sen \etal \cite{sen2012robust} propose a method based on a patch-matching algorithm for HDR reconstruction.
Hu \etal\cite{hu2013hdr} propose a displacement estimation method which converts images by the intensity mapping function and then merging images in the transformed domain for HDR image generation, which implicitly align LDR images by searching and aggregating similar patches. Hafner \etal \cite{hafner2014simultaneous} propose a method to jointly estimate the optical flow and reconstruct HDR image. However, since the alignment process in the image domain is vulnerable to large motion and excessively dark or bright regions, these methods tend to generate artifacts in the aligned images.
%And these methods employ a simple method for merging aligned LDR images, which cannot effectively eliminate those artifacts. 

\subsection{CNN Based Methods}
As with other computer vision tasks, CNNs have been applied to HDR imaging. Eilertsen \etal \cite{eilertsen2017hdr} propose an encoder-decoder network to generate an HDR image from a single LDR image. Endo \etal \cite{endoSA2017} synthesize multiple LDR images with different exposures from a single LDR image by CNNs, and then merge them to reconstruct an HDR image. These single-image-based methods are unable to reconstruct the textures on saturated regions accurately. 

To generating more accurate images, more attention is paid on obtaining HDR images from a sequence of LDR images captured with different exposures. Kalantari \etal \cite{kalantari2017deep} propose the first CNN-based method for HDR imaging, where the input LDR images are first aligned by optical flow and then the aligned LDR images are fed to CNNs to reconstruct an HDR image.
% Wu \etal \cite{wu2018deep} employ a U-net structure to reconstruct an HDR image without explicit alignment of the input images.
% Wu \etal \cite{wu2018deep} reconstruct an HDR image by using a U-net type structure and directly concatenating the features from input images but without explicit alignment of the input images. 
In stead of using explicit alignment, Wu \etal \cite{wu2018deep} directly concantenated the features extracted from input LDR images and forwarded them to a deep model with the U-net structure to reconstruct HDR images.
Yan \etal \cite{yan2020deep} introduce a non-local structure \cite{wang2018non} into the U-net as for implicit alignment. Yan \etal \cite{yan2019attention} propose an attention module to learn to identify misaligned elements before merging the LDR images.
% Pu \etal \cite{pu2020robust} achieve pyramidal alignment by using the deformable convolution \cite{zhu2019deformable} in multi-scale features and reconstruct an HDR image from these aligned features. 
Pu \etal applied the deformable convolution \cite{zhu2019deformable} to multi-scale features, which aligned LDR images in a pyramidal manner, and reconstructed the corresponding HDR images.
To reduce the computational cost by CNN-based methods, Prabhakar \etal \cite{prabhakar2020towards} propose an efficient method that performs all operations in low resolution and upscales the result to the required full resolution.
%Inspired by non-local structure \cite{wang2018non}, Choi \etal \cite{choi2020pyramid} tried to solve the problem of ghosting artifacts by calculating the inter-similarity between LDR images for every pixel, which are used to align non-reference features toward the reference feature. 
Similar to our work, a few studies consider using optical flow for the alignment. But they either use a pre-trained estimator \cite{prabhakar2019fast} or optimize the estimator through the reconstruction loss \cite{kalantari2019deep}, which may lead to suboptimal results.

\begin{figure}[htbp]
\begin{center}
%\fbox{\rule{0pt}{2in} \rule{.9\linewidth}{0pt}}
\includegraphics[width=0.9\linewidth]{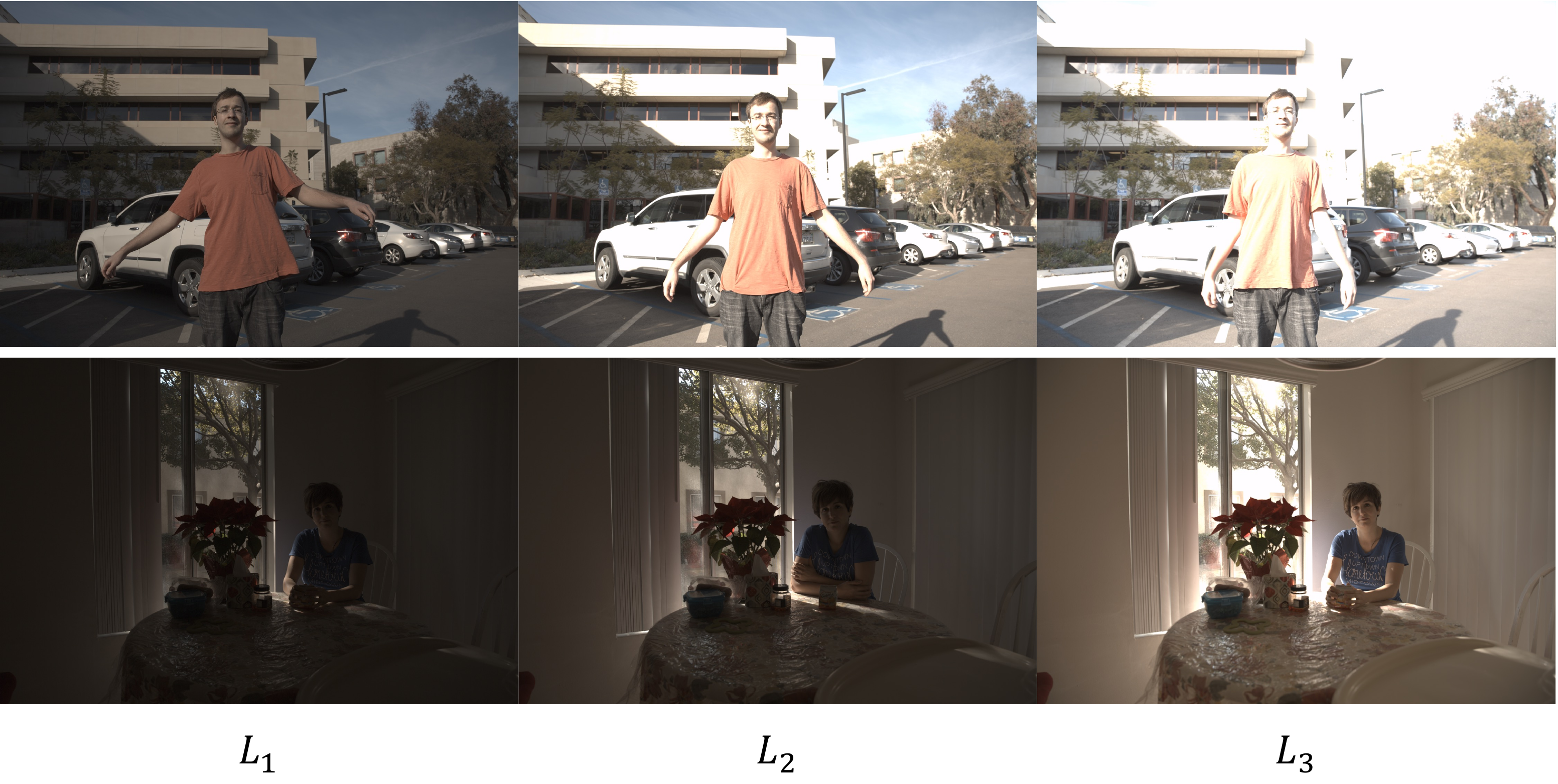}
\end{center}
  \vspace*{-4mm}
  \caption{Three LDR images of the same scene captured with three different exposures. $L_{1}$, $L_{2}$, and $L_{3}$ denote the images captured with the low, medium, and high exposure, respectively.}
\label{fig:multi-exp}
\end{figure}
%%%%%%%%%%%%%%%%%%%%%%%%%%%%%%%%%%%%%%%%%%%%%%%%%%%%%%%%%%%%%%%%%%%%%%%%%%%%%%%%%%%
\begin{figure*}[htbp]
\begin{center}
%\fbox{\rule{0pt}{2in} \rule{.9\linewidth}{0pt}}
\includegraphics[width=\linewidth]{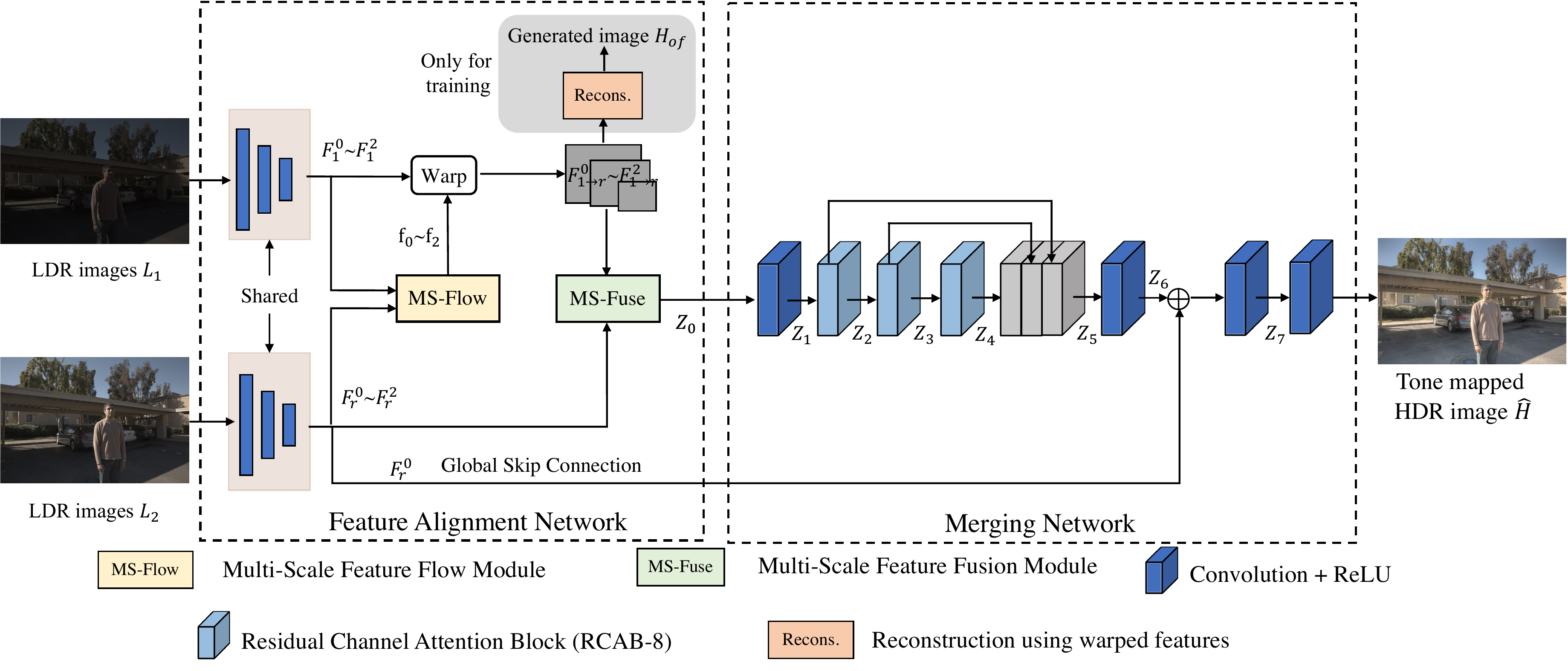}
\end{center}
  \vspace*{-4mm}
   \caption{Overview of the proposed network. It consists of two sub-networks: feature alignment and merging networks. The alignment network warps the features of the non-reference images onto those of the reference image using optical flow. The merging network takes the warped features as input and reconstructs an HDR image.}
\label{fig:arch}
\end{figure*}

%------------------------------------------------------------------------- 
\section{Proposed Method}
Given a series of LDR images, $L_1, L_2,..., L_k$, captured with different exposures, the goal of HDR imaging is to generate an HDR image $H$ corresponding to a selected reference image $L_r$.

There are two samples with different exposure settings shown in Fig.~\ref{fig:multi-exp}. The sample in the first row shows that $L_3$ has little effect for image restoration of $L_2$ since there are large areas of over-exposure region in $L_3$. While the sample in the second row shows that $L_3$ can be helpful for image restoration of $L_2$. In this case, the model can easily produce high-quality HDR images duo to the efficient information. Without the input of $L_3$, the model can also generate a high-quality HDR image though the model needs to be more effective.

Unlike the previous methods taking three LDR images as input, we use two LDR images, $L_1$ and $L_2$ as inputs, sorted in the order of exposures and set $L_2$ as the reference image by considering the properties in $L_3$. And two images for input can also reduce computational costs. 

Following the settings of previous studies \cite{kalantari2017deep,yan2019attention}, we first map the LDR images into the HDR domain using gamma correction and then feed them into the network.
% As described in \cite{kalantari2017deep}, the images in LDR domain are helpful for detecting the noisy or saturated regions whereas the images in HDR domain help identify the differences from the reference image. Thus, we first map the LDR images into the HDR domain using gamma correction before feeding them into the network \cite{kalantari2017deep}.
To be specific, we map $L_i$ to $H_i$ by
\begin{equation}
    H_{i} = L_{i}^{\gamma}/t_i, \;\;
    % \forall	
    i=1, 2, 
\end{equation}
where $\gamma$ denotes the gamma correction parameter and followed \cite{poynton2012digital} we use $\gamma=2.2$ in this paper. $t_i$ is the exposure time of $L_i$.  
As the suggestion in \cite{kalantari2017deep}, we concatenate $L_i$ and $H_i$ in the channel dimension to obtain a six-channel tensor $X_i=[L_i, H_i], i=1,2$, and input $X_1$ and $X_2$ to the network.

Our network consists of two sub-networks, the alignment network and merging network, as shown in Fig.\ref{fig:arch}. We first describe the alignment network (Sec.\ref{sec:alignment}) and then explain the merging network (Sec.\ref{sec:merging}).

%%%%%%%%%%%%%%%%%%%%%%%%%%%%%%%%%%%%%%%%%%%%%%%%%%%%%%%%%%%%%%%%%%%%%%%%%%%%%%%%%%%

%------------------------------------------------------------------------- 
\subsection{Feature Alignment Network} \label{sec:alignment}
%The feature alignment network first extracts multi-scale features $F_{i}^{s}$ $(s=1,2,3)$ from each of the two input LDR images $(i=1,2)$. Considering the computational efficiency, we use three scales.
The feature alignment network first extracts multi-scale feature maps from the input tensor $X_i$. Specifically, the feature extractor consists of a convolution layer with stride $=1$ and the following two layers with stride $=2$, which forms multi-scale feature maps with scale $= 0,1,$ and $2$.
We represent the feature map at scale $s\in \{0,1,2\}$ as $F_{i}^{s} \in \mathbb{R}^{H_{s}\times W_{s}\times C_{s}}$, where $H_{s} = H/2^{s}$, $W_{s}=H/2^{s}$, and $C_{s}=C$. These feature maps will be used in the subsequent modules. 
%They all output the same number of feature maps.
%{\color{blue} 
For clarity, 
% we define notations $X_r$ and $F_{r}^{3}$ to indicate $X_2$ and $F_{i}^{3}$ corresponding to the reference LDR image in some special context. 
we use the index $r(=2)$ to indicate the reference LDR images; thus, $X_r=X_2$ and $F_r^s=F_2^s$ in what follows.

\begin{figure}[t]
\begin{center}
%\fbox{\rule{0pt}{2in} \rule{.9\linewidth}{0pt}}
\includegraphics[width=0.9\linewidth]{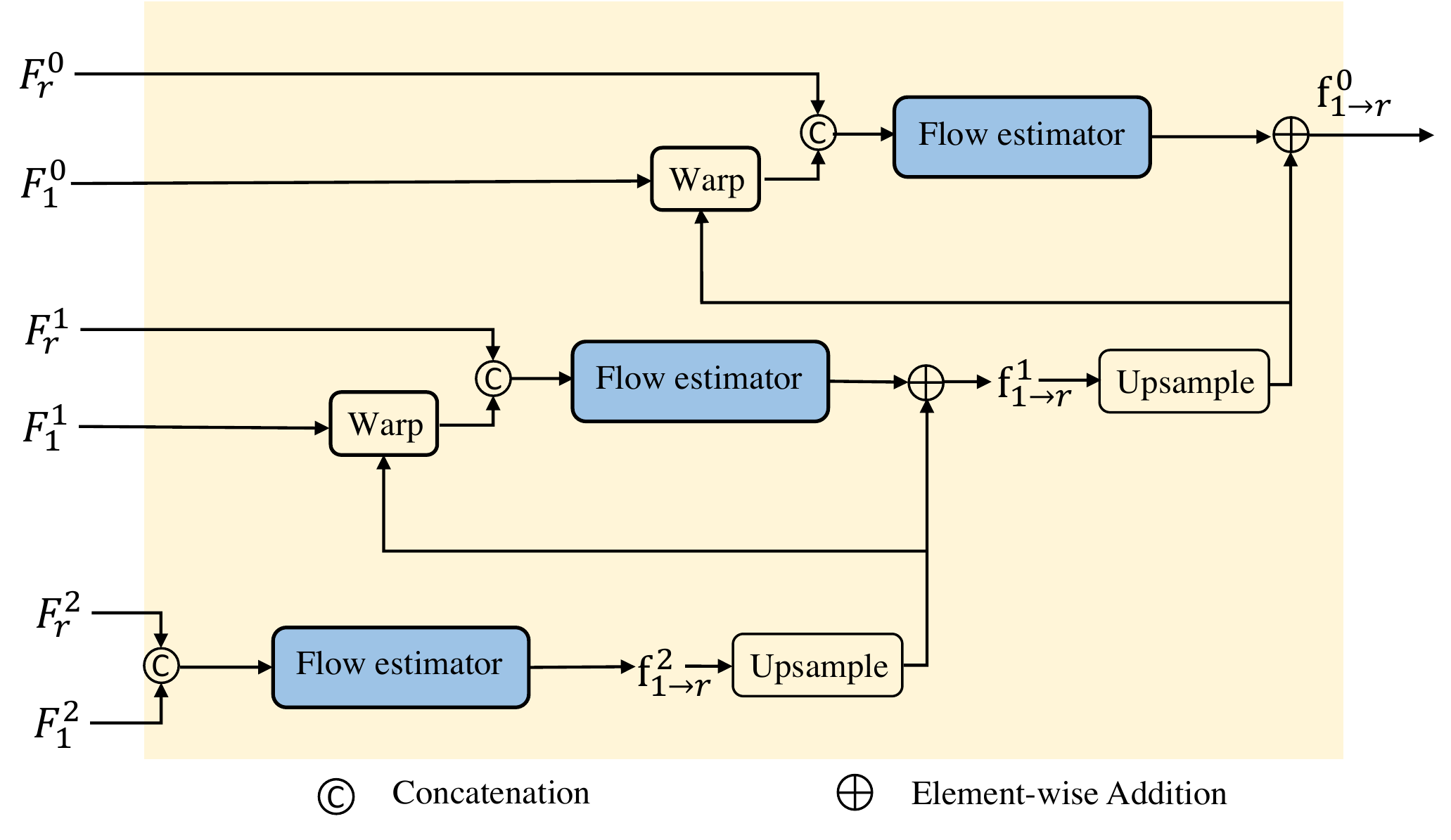}
\end{center}
  \vspace*{-4mm}
   \caption{Architecture of the multi-scale feature flow module (MS-Flow). It follows the coarse-fine manner to generates multi-scale optical flows and the multi-scale feature maps aligned to the reference image.}
\label{fig:msflow}
\end{figure}

%%%%%%%%%%%%%%%%%%%%%%%%%%%%%%%%%%%%%%%%%%%%%%%%%%%%%%%%%%%%%%%%%%%%%%%%%%%%%%%%%%%

%------------------------------------------------------------------------- 
\subsubsection{Multi-Scale Feature Flow Module (MS-Flow)}
%Although standard methods for optical flow estimation~\cite{ranjan2017optical,sun2018pwc} take RGB images as input and estimate optical flow, it tends to be vulnerable to illumination changes. Thus, we estimate the optical flow in feature space to improve the robustness.
%Thus, we propose to use the feature maps for HDR reconstruction as input to improve the robustness of flow estimation. 
% The architecture of multi-scale feature flow estimation module is shown in Figure~\ref{fig:msflow}. 
Following SPyNet~\cite{ranjan2017optical} and PWC-Net~\cite{sun2018pwc}, we estimate the optical flow in a coarse-to-fine manner, as shown in Fig.~\ref{fig:msflow}.
%Inspired by SPyNet~\cite{ranjan2017optical} and PWC-Net~\cite{sun2018pwc}, we propose a multi-scale feature flow module for HDR imaging based on hierarchical coarse-to-fine feature warping.
The estimated flows at the coarser scales can capture the large motions. On the other hand, the flows at the finer scales will be helpful to capture small motions. %But the difference is that we directly use the features for HDR image reconstruction as the input for flow estimation.
%The estimated flows at the coarser scales can capture the large motions, on the other hand, the flows at the finer scales can capture and are used to initialize the inputs for the CNN in the finer scales, which are responsible for estimating the smaller motions.

We first concatenate the coarsest scale features $F_{i}^{s}$ with $F_r^s$ in the channel dimension and feed it to a flow estimator. The estimator consists of five convolution layers with $7\times7$ kernel size and generates the $s$-th scale optical flow $\mathbf{f}_{1\rightarrow r}^s$. Then, we upsample $\mathbf{f}_{1\rightarrow r}^s$ by factor $=2$ and use it to warp the non-reference feature map $F_{1}^{s-1}$ onto the reference feature map $F_{r}^{s-1}$.
Specifically, we map each pixel $\mathbf{p}_1^{s-1}$ in $F_{1}^{s-1}$ to its estimated correspondence in $F_{r}^{s-1}$ as 
%Specifically, when the flow warps each point $\mathbf{p}_1^{s-1}$ in $F_{1}^{s-1}$ to $\mathbf{p}_r^{s-1}$ in $F_{r}^{s-1}$, this is described by
\begin{equation}
\mathbf{p}_r^{s-1} = \mathbf{p}_1^{s-1} + \mathbf{\tilde{f}}_{1\rightarrow r}^{s}(\mathbf{p}_1^{s-1}),
\label{eq:warp}
\end{equation}
where $\mathbf{\tilde{f}}_{1\rightarrow r}^{s}$ represents the upsampled flow of the $s$-th scale flow $\mathbf{f}_{1\rightarrow r}^s$.
We then concatenate the warped and reference feature maps in the channel dimension and feed it to the subsequent flow estimator. The output of the flow estimator is then element-wise added to the upsampled flow, yielding the flow at scale $s-1$. We iterate this procedure for $s=1$ and $0$, obtaining multi-scale optical flows $\mathbf{f}_{1\rightarrow r}^s$ and the warped multi-scale feature maps $F_{i\rightarrow r}^{s}$.

%------------------------------------------------------------------------- 
\subsubsection{Multi-Scale Feature Fusion Module (MS-Fuse)}
\begin{figure*}[t]
  \begin{minipage}[t]{0.5\linewidth}
    \centering
    \includegraphics[width=0.95\linewidth]{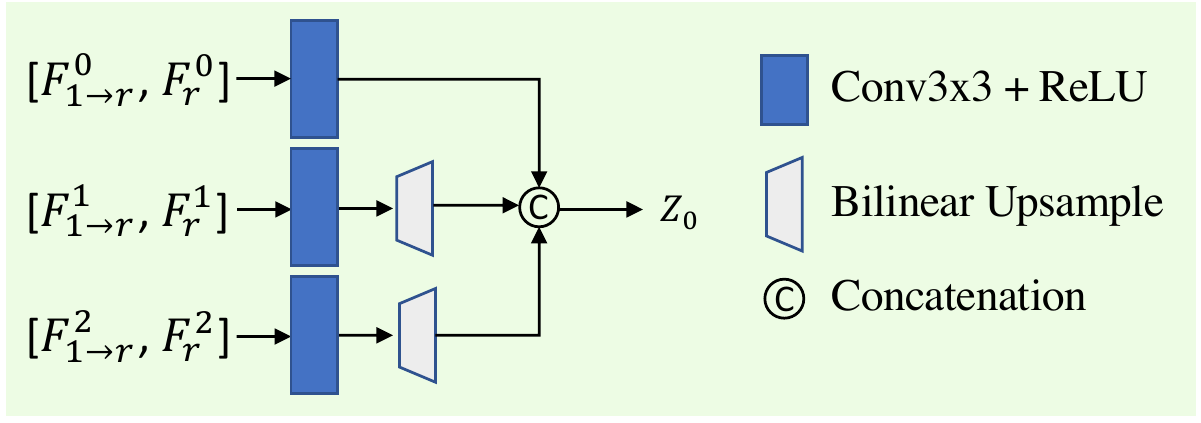}
    \caption{Multi-scale feature fusion.}
    \label{fig:msff}
  \end{minipage}%
  \begin{minipage}[t]{0.5\linewidth}
    \centering
    \includegraphics[width=0.95\linewidth]{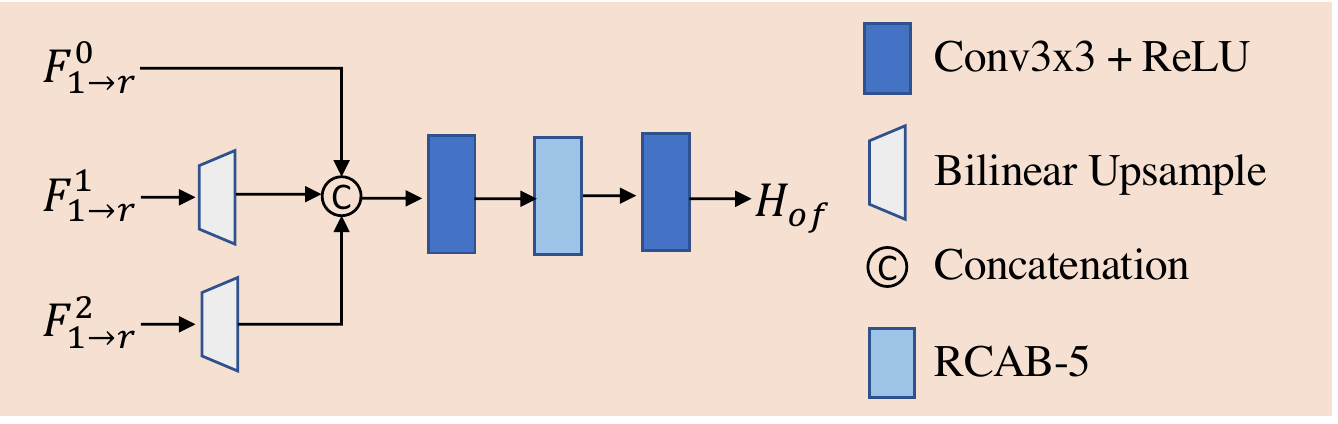}
    \caption{Reconstruction of $H_{of}$.}
    \label{fig:reg_rec}
  \end{minipage}
\end{figure*}

As shown in Fig.~\ref{fig:msff}, the multi-scale feature fusion module takes the concatenated feature maps at each scale $\bar{F}^s = [F_{1\rightarrow r}^s, F_r^s]$. We apply a convolution layer with the kernel size of $3\times 3$ followed by ReLU to the finest feature map $\bar{F}^0$ to obtain a feature map $O^0$. For the feature maps $\bar{F}^1$ and $\bar{F}^2$, we first apply a convolution layer with the same kernel size and then upsample the outputs with bilinear interpolation so that the resulting maps become the same size as the finest one. Finally, all the outputs are concatenated as $Z_0 = [O^0, O^1, O^2]$ and then used as input for the following merging network.

%------------------------------------------------------------------------- 

% 
%%%%%%%%%%%%%%%%%%%%%%%%%%%%%%%%%%%%%%%%%%%%%%%%%%%%%%%%%%%%%%%%%%%%%%%%%%%%%%%%%%%
\subsubsection{Reconstruction using Warped Features}
\begin{figure}[t]
\begin{center}
\includegraphics[width=.95\linewidth]{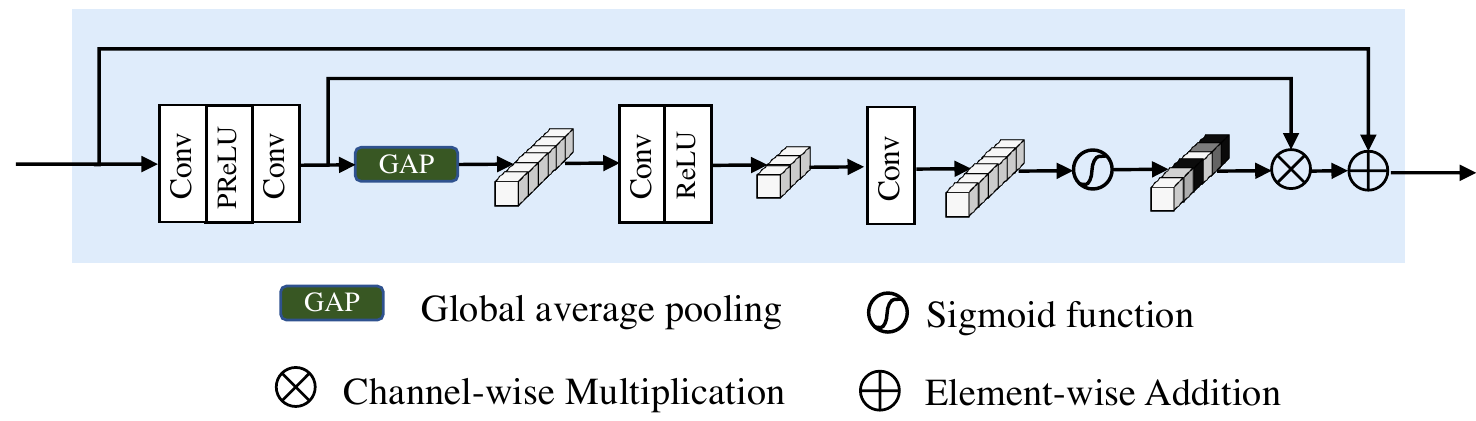}
\end{center}
\vspace*{-4mm}
\caption{Architecture of a residual channel attention block (RCAB) \cite{zhang2018image}.}
\label{fig:cab}
\end{figure}

Unlike the previous HDR imaging studies using optical flow, we reconstruct an HDR image $H_{of}$ using the feature maps $F_{1\rightarrow r}^s$, which are the feature maps right after warping by the optical flow. 
Our intention behind this reconstruction is to directly guide the network to generate accurate optical flow and perform better alignment.
%We expect this to make the optical flow more accurate.
As shown in Fig.~\ref{fig:reg_rec}, we first upsample the feature maps $F_{1\rightarrow r}^1, F_{1\rightarrow r}^2$ so as to be the same size as the finest feature map $F_{1\rightarrow r}^0$. We then concatenate them and feed them to a series of convolution layers with the kernel size of $3\times 3$ followed by ReLU and five residual channel attention blocks (RCAB) \cite{zhang2018image}; see Fig.~\ref{fig:cab} for the detail of the RCAB. We then calculate $\ell_1$ loss between the reconstructed HDR image $H_{of}$ and its ground truth HDR image, as will be explained later.

%===========================================================
%%%%%%%%%%%%%%%%%%%%%%%%%%%%%%%%%%%%%%%%%%%%%%%%%%%%%%%%%%%%%%%%%%%%%%%%%%%%%%%%%%%
\subsection{Merging Network}  \label{sec:merging}
Following the previous methods \cite{zhang2018image,yan2019attention}, we employ an attention mechanism to merge the feature maps and generate an HDR image; in specific, we use the RCAB. As shown in Fig.~\ref{fig:arch}, the merging network takes the concatenated feature maps $Z_0 = [O^0, O^1, O^2]$. We apply a convolution layer and three RCABs to $Z_0$ and then concatenate the outputs of each RCAB as $Z_5 = [Z_2, Z_3, Z_4]$. Applying three convolutions and a global skip connection with $F_r^0$, the merging network outputs a final HDR image. 
%As shown in Figure~\ref{fig:arch}, the merging network takes the concatenated features $Z_0$ as input. $Z_1$ is obtained after a convolution layer, then feature maps $Z_2, Z_3, Z_4$ are generated by feeding $Z_1$ into 3 RCAB*8 (shown in Figure~\ref{fig:cab}). By applying three convolutional layers with global skip connection and a sigmoid function on the concatenated features $Z_5$, the merging network outputs an HDR image. 

%%%%%%%%%%%%%%%%%%%%%%%%%%%%%%%%%%%%%%%%%%%%%%%%%%%%%%%%%%%%%%%%%%%%%%%%%%%%%%%%%

\subsection{Loss Function}
Following \cite{kalantari2017deep}, we consider the optimization in the domain of tonemapped HDR images because the HDR images are usually displayed after tonemapping and training the network on the domain is more effective than that on the original domain of HDR images. Thus, we employ the $\mu$-law for tone mapping as suggested in \cite{kalantari2017deep}, which is formulated as
\begin{equation}
    T(H) = \frac{log(1+\mu H)}{log(1+\mu)} ,
\end{equation}
where $\mu$ is set to $5,000$ throughout our experiments.
It is also reported in \cite{zhao2016loss,yan2019attention} that minimizing the $L_1$ norm between the predicted HDR image $\hat{H}$ and its ground truth $H$ in the tone-mapped domain works better than others.
Following their studies, we use the following $\ell_1$ loss,
\begin{equation}
    L_{tm}=\lVert T(\hat{H})-T(H) \rVert_1. %+ \lambda \lVert T(H_{of})-T(H) \lVert_1.
    \label{loss-tm}
\end{equation}

For the standard optical flow estimators such as SPyNet~\cite{ranjan2017optical} and PWC-Net~\cite{sun2018pwc}, they are trained on the datasets with the standard exposure settings (e.g. Sintel \cite{butler2012naturalistic}, KITTI \cite{geiger2012we}, and Middlebury \cite{scharstein2014high}). However, there is no dataset containing ground truths of optical flow maps for the HDR imaging task. Inspired by the photometric loss \cite{yu2016back} for self-supervised learning of optical flow, we use $\ell_1$ loss between the reconstruction $H_{of}$ and its ground truth $H$ in the tone-mapped domain to provide supervision for the optical flow learning,
\begin{equation}
    L_{reg}= \lVert T(H_{of})-T(H) \lVert_1.
    \label{loss-reg}
\end{equation}

The reconstruction of $H_{of}$ is only based on the warped features from the non-reference images. Then $L_{reg}$ computed on $H_{of}$ and the ground truth imposes a heavy constraint on these warped features to provide more accurate gradients to the warping field than the $L_{tm}$ which involves both the warped features and the features from the reference image.

Our total loss is taken as the weighted sum of two losses
\begin{equation}
    L=  L_{tm} +  \lambda L_{reg},
    \label{loss-total}
\end{equation}
where we use $\lambda=2$ in this paper.

%%%%%%%%%%%%%%%%%%%%%%%%%%%%%%%%%%%%%%%%%%%%%%%%%%%%%%%%%%%%%%%%%%%%%%%%%%%%%%%%%%%
%===========================================================
\begin{table*}[t]
\begin{center}
\caption{Quantitative comparison on the Kalantari’s test sets \cite{kalantari2017deep}. The numbers in the table are the average values of the 15 test images.}
\label{Table: main} \small
\setlength{\tabcolsep}{5pt}
\begin{tabular}{l|ccccc}
\hline
Methods & PSNR-$\mu$ &  PSNR-L & SSIM-$\mu$ &{SSIM-L} & HDR-VDP-2\\
\hline
TMO \cite{endoSA2017}  & 8.3120& {8.8459}  & 0.5029 &{0.0924} & 44.3345\\
HDRCNN \cite{eilertsen2017hdr} & 13.7054 & {13.8956} & 0.5924  & {0.3456}& 47.5690\\
Sen \cite{sen2012robust} & 40.9689 & {38.3425} & 0.9859 &{0.9764} & 60.3463 \\
Kalantari \cite{kalantari2017deep} & 42.7177 & {41.2200}  & 0.9889 & {0.9829}& 61.3139 \\
Wu \cite{wu2018deep}    & 41.9977 & {\underline{41.6593}} & 0.9878& {0.9860} & 61.7981 \\
AHDR \cite{yan2019attention} & 43.7013  & {41.1782} & 0.9905 & {0.9857} & 62.0521 \\
PANet \cite{pu2020robust}      & 43.8487 & {41.6452} & {0.9906}  & {\underline{0.9870}}& 62.5495 \\
PSFNet \cite{ye2021progressive}         & \underline{44.0613} & {41.5736} &  \underline{0.9907} & {0.9867}& \textbf{63.1550} \\
Ours           & \textbf{44.3298}& {\textbf{41.8936}}& {\textbf{0.9911}} &  \textbf{0.9878}  & \underline{63.1190} \\ 
\hline
\end{tabular}
\end{center}
\vspace*{-4mm}
\end{table*}

\begin{figure*}[tb] \centering
    \includegraphics[width=0.485\textwidth]{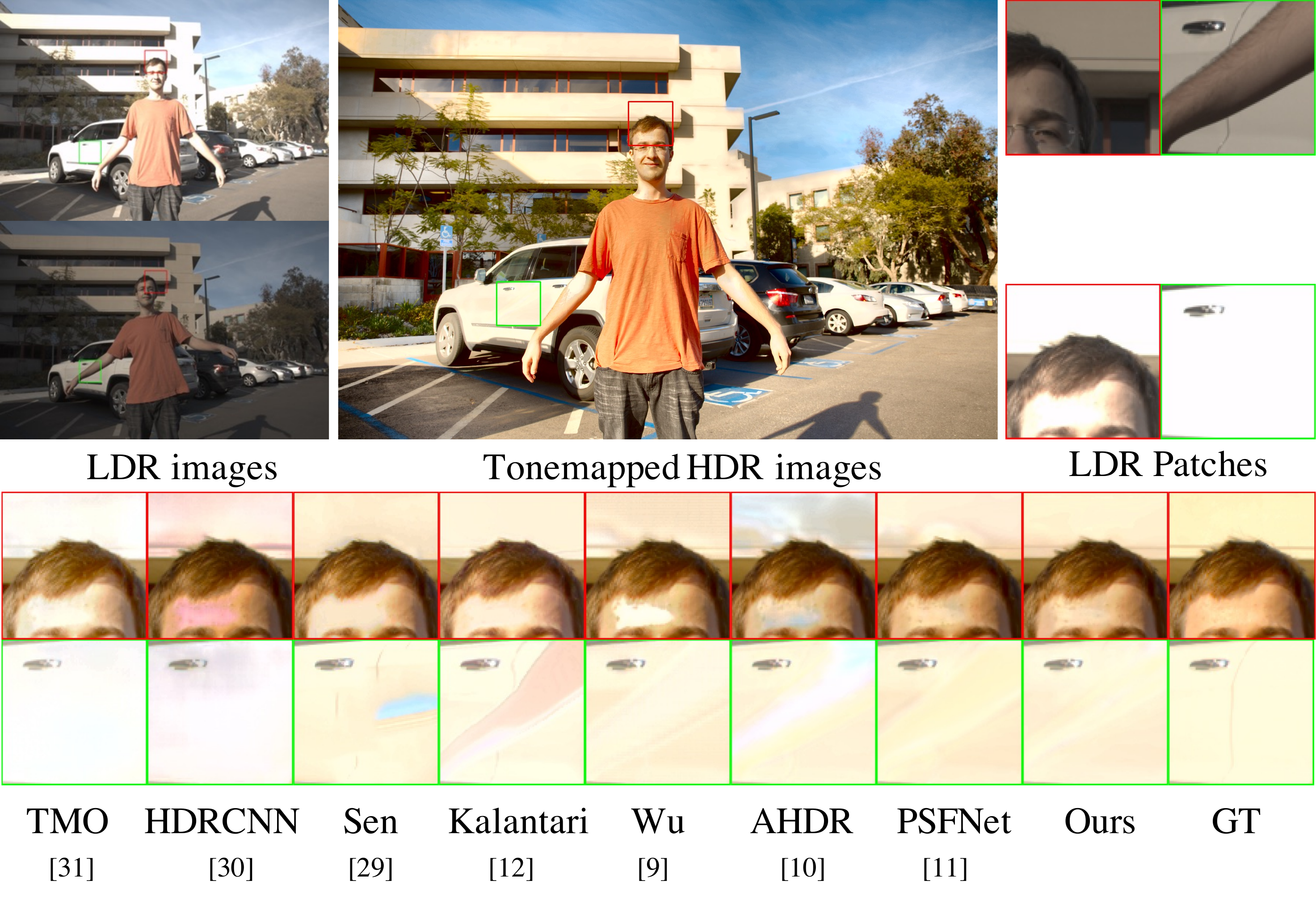}
    \includegraphics[width=0.485\textwidth]{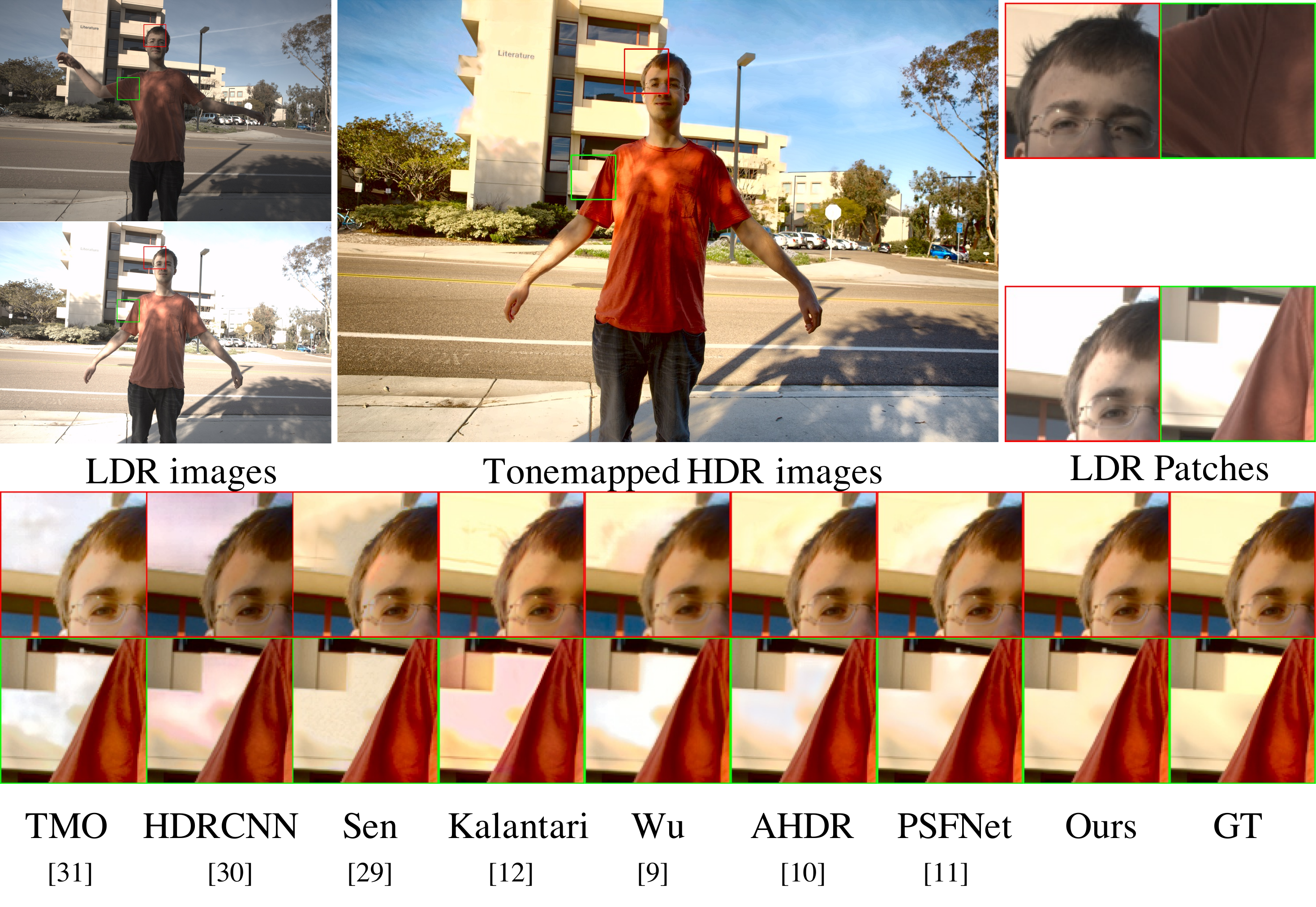}
    \\
    \makebox[0.485\textwidth]{\scriptsize Testing data (08) in \cite{kalantari2017deep}}
    \makebox[0.485\textwidth]{\scriptsize Testing data (09) in \cite{kalantari2017deep}}
    \\
    \caption{Results from the test set of \cite{kalantari2017deep}. Upper row from left to right: the two input LDR images, the HDR image produced by the proposed method, and (zoomed-in) LDR image patches with two identical positions/sizes (in green and red). Lower row: the same patches of the HDR images produced by different existing methods.}
\label{fig:main}
\end{figure*}

%--------------------------------------

\begin{figure*}[tb] \centering
    \includegraphics[width=0.485\textwidth]{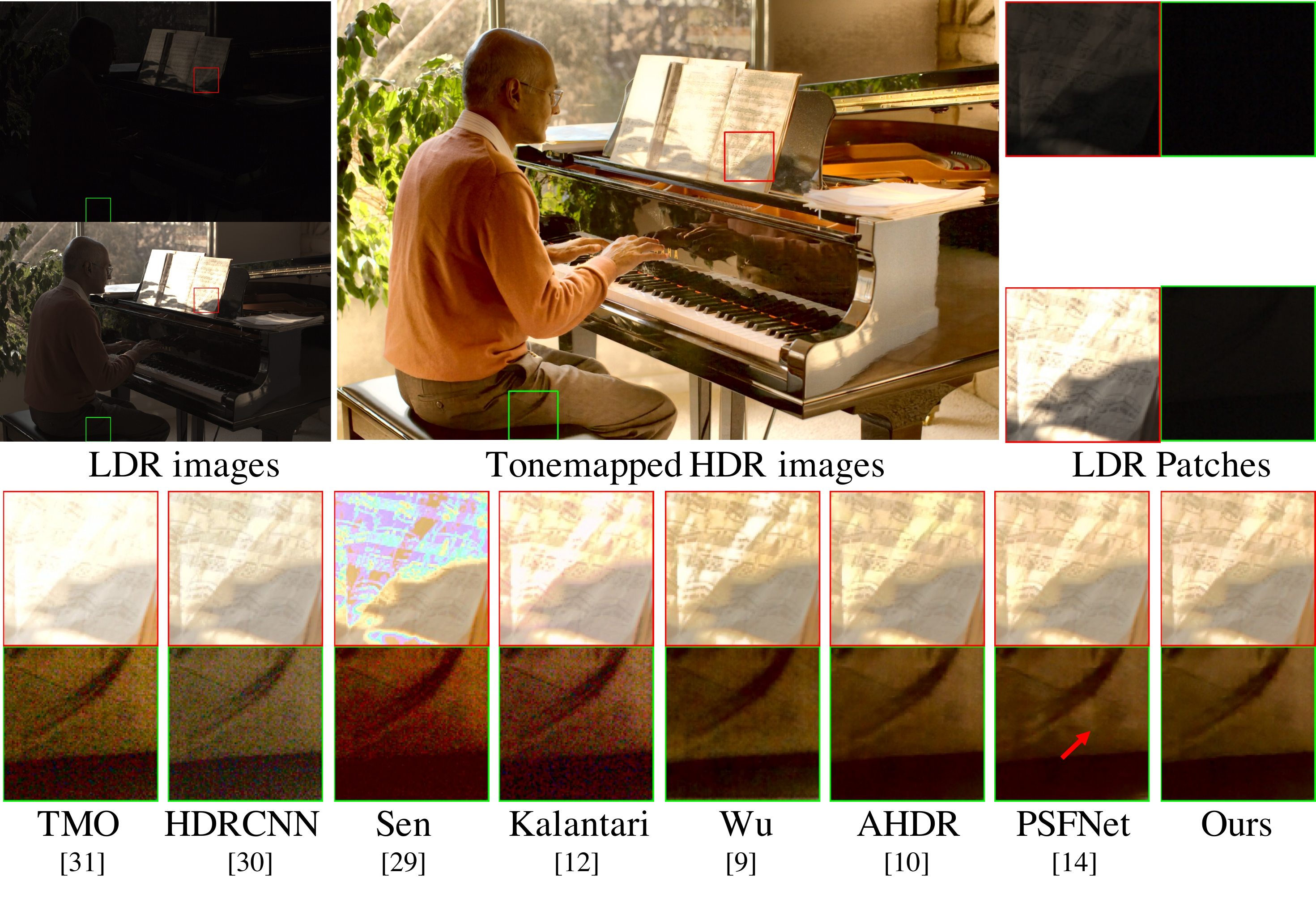}
    \includegraphics[width=0.485\textwidth]{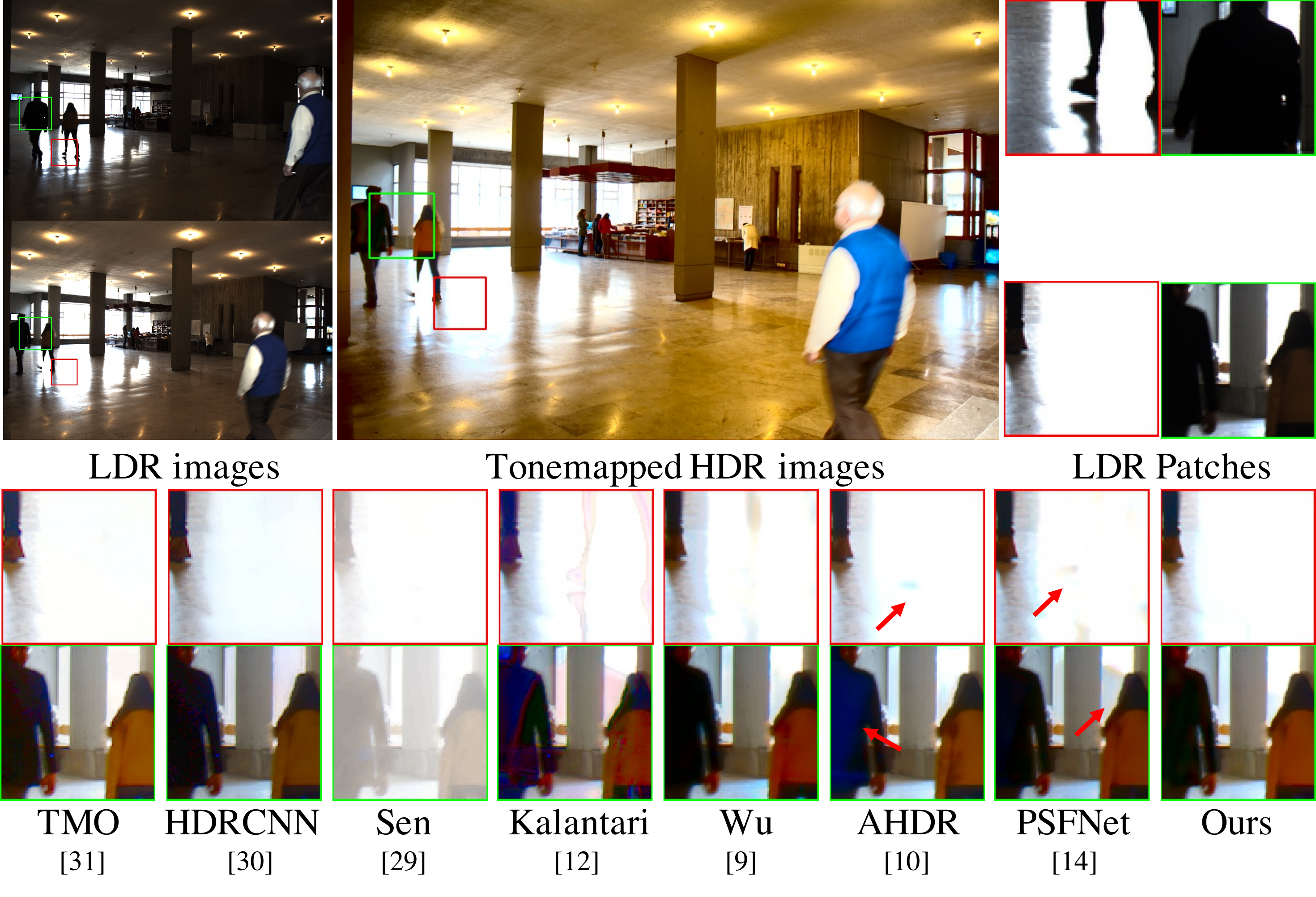}
    \\
    \makebox[0.485\textwidth]{\scriptsize Example of Sen \etal’s dataset \cite{sen2012robust}}
    \makebox[0.485\textwidth]{\scriptsize Example of Tursun \etal’s dataset \cite{tursun2016objective}}
    \\
    \caption{Visual comparisons on the datasets without ground truth. See Fig.~\ref{fig:main} for the explanation of the panels.}
    \label{fig:noref}
\end{figure*}

\section{Experiments}
\subsection{Experimental Settings}
\subsubsection{Training Data}  
To train our network, we adopt the HDR dataset \cite{kalantari2017deep} which consists of 74 samples for training and 15 samples for testing. We use the former for training our model. Each sample includes a ground truth HDR image and three LDR images with different exposure settings of $\{-2, 0, +2\}$ or $\{-3, 0, +3\}$. All the images are resized to the resolution of $1000\times 1500$.

%%%%%%%%%%%%%%%%%%%%%%%%%%%%%%%%%%%%%%%%%%%%%%%%%%%%%%%%%%%%%%%%%%%%%%%%%%%%%%%%%%%
\subsubsection{Testing Data} 
Following recent studies, we choose the following datasets for testing.
We evaluate our method on the 15 scenes of the dataset of \cite{kalantari2017deep}, where we perform a quantitative evaluation using the provided ground truths. We also test the proposed method on the datasets of Sen \etal \cite{sen2012robust} and Tursun \etal \cite{tursun2016objective}. Since these datasets do not contain ground truths of HDR images, we compare the reconstructed HDR images by our method with those by state-of-the-art methods for qualitative evaluation. 

%%%%%%%%%%%%%%%%%%%%%%%%%%%%%%%%%%%%%%%%%%%%%%%%%%%%%%%%%%%%%%%%%%%%%%%%%%%%%%%%%%%
\subsubsection{Evaluation Metrics} \label{sec:metrics}
As used in the existing studies, we use PSNR-$\mu$ and SSIM-$\mu$ for primary metrics, which are PSNR and SSIM values in the tone-mapped domain using $\mu$-law.
%It is argued in \cite{kalantari2017deep} that HDR imaging methods should be evaluated in the tone-mapped domain, as we usually use generated HDR images after tone-mapping. Following \cite{kalantari2017deep}, we use PSNR-$\mu$ and SSIM-$\mu$ for primary metrics, which are PSNR and SSIM values in the tone-mapped domain achieved by the $\mu$-law.
We show PSNR and SSIM values in the linear domain, which are denoted by PSNR-L and SSIM-L for completeness. We also report HDR-VDP-2 \cite{mantiuk2011hdr}, which is designed to evaluate the quality of HDR images.

%%%%%%%%%%%%%%%%%%%%%%%%%%%%%%%%%%%%%%%%%%%%%%%%%%%%%%%%%%%%%%%%%%%%%%%%%%%%%%%%%%%
\subsection{Implementation Details}
For training, we first crop the training images into patches of $256\times256$ pixel size with a stride of 128 pixels. %The crop is conducted on both LDR images and the corresponding HDR ground truth.
We then apply random rotation and flipping for data augmentation to avoid over-fitting. We use the Adam optimizer \cite{KingmaB14} with $\beta_1=0.9$, $\beta_2=0.999$, initial learning rate $1\times10^{-4}$ and set the batch size to 8. We train our model for 210 epochs and employ the cosine annealing strategy \cite{loshchilov2016sgdr} to steadily decrease the learning rate from an initial value to $1\times10^{-6}$. We implement our model using PyTorch \cite{paszke2017automatic} on NVIDIA GeForce RTX 2080 GPUs.

%%%%%%%%%%%%%%%%%%%%%%%%%%%%%%%%%%%%%%%%%%%%%%%%%%%%%%%%%%%%%%%%%%%%%%%%%%%%%%%%%%%
\subsection{Comparison with the State-of-Art Methods}
%We use Kalantari’s testset \cite{kalantari2017deep} and the above datasets without ground truths \cite{sen2012robust,tursun2016objective} in the experiments.
We compare the proposed method with existing methods. Specifically, we compare our model with two HDR imaging methods based on a single LDR image, TMO \cite{endoSA2017} and HDRCNN \cite{eilertsen2017hdr}, and five HDR imaging methods based on multi LDR images, the patch-based method \cite{sen2012robust}, the flow-based method with CNN merger \cite{kalantari2017deep}, the U-net structure without optical flow \cite{wu2018deep}, the attention-guide method (AHDR) \cite{yan2019attention}, pyramidal alignment network (PANet) \cite{pu2020robust}, and progressive and selective fusion network (PSFNet) \cite{ye2021progressive}. For all the methods, we used the authors' code for comparison, except for \cite{pu2020robust} since their code is not available as of the time of writing this paper.

\subsubsection{Evaluation on Kalantari \etal’s Dataset}
Figure \ref{fig:main} shows two examples on the test set of \cite{kalantari2017deep}; see the supplementary material for more visualization results. The input LDR images contain saturated background and foreground motions. We can observe from the results of the single-image methods, TMO \cite{endoSA2017} and HDRCNN \cite{eilertsen2017hdr}, that they cannot sufficiently recover the detailed textures and generate artifacts in the over-exposed regions; they also suffer from the color distortion. The patch-based method of Sen \etal \cite{sen2012robust} generates some artifacts due to the failure of finding patches correctly. Kalantari \etal's method \cite{kalantari2017deep} cannot completely eliminate the effects of the occlusion. The method of Wu \etal \cite{wu2018deep} cannot deal with over-exposed regions and then produces artifacts on motion areas. Non-aligned methods (\ie AHDR \cite{yan2019attention} and PSFNet \cite{ye2021progressive}) yield artifacts in the saturated areas and also suffer from ghosting artifacts due to the large motions. %produces the best qualitative results; it produces less color distortion and recovers the textures more accurately.
Compared with them, our proposed method produces less color distortion and recovers the textures more accurately, leading to the best qualitative results.

Table \ref{Table: main} shows the quantitative evaluation on the same dataset. In specific, we report the averaged values over 15 test scenes. It can be seen that the proposed method achieves better performance than the others in terms of PSNR-$\mu$, SSIM-$\mu$, PSNR-L and SSIM-L. Also, our method achieves comparable performance to the state-of-the-art method \cite{ye2021progressive} in terms of the HDR-VDP-2 metric.

%------------------------------------------------------------------------- 
%%%%%%%%%%%%%%%%%%%%%%%%%%%%%%%%%%%%%%%%%%%%%%%%%%%%%%%%%%%%%%%%%%%%%%%%%%%%%%%%%%%
\subsubsection{Evaluation on Datasets w/o Ground Truth}
We also provide comparisons using Sen’s \cite{sen2012robust} and Tursun’s \cite{tursun2016objective} datasets. These datasets do not have ground truths of HDR images and thus we qualitatively compare the generated HDR images.

Some examples of the results are shown in Fig.~\ref{fig:noref}. The single image methods, TMO \cite{endoSA2017} and HDRCNN \cite{eilertsen2017hdr},  generate serious noises and color distortions in the under-exposed regions. The patch-based method (Sen et al \cite{sen2012robust}) also generates severe artifacts. The method of Kalantari \etal \cite{kalantari2017deep} produces artifacts due to the alignment error and also generates serious noises in the under-exposed regions. These are arguable because of the misalignment by the estimated optical flow and the limitation of the merging method. Wu \etal’s method \cite{wu2018deep} tends to yield over-smoothness and generate ghosting artifacts on the large motion areas. AHDR \cite{yan2019attention} yields color distortions and also suffers from ghosting artifacts due to large motions shown in Fig.~\ref{fig:noref} (b). PSFNet \cite{ye2021progressive} generates ghosting artifacts in the motion regions and generates the geometric distortions shown in Fig.~\ref{fig:noref} (a). On the other hand, our method produces better results with noticeably reduced geometric and color distortions compared with others.

%%%%%%%%%%%%%%%%%%%%%%%%%%%%%%%%%%%%%%%%%%%%%%%%%%%%%%%%%%%%%%%%%%%%%%%%%%%%%%%%%%%
\begin{table*}[htbp]
\begin{center}
\caption{Results of ablation tests on Kalantari’s test set. The upper row shows the effects of channel attention (CA), multi-scale feature flow module (MS-Flow), feature flow module (FF), and multi-scale feature fusion module (MS-Fuse). The lower row shows the effects of the choice of optical flow.}
\label{Table:ablation} \small
\begin{tabular}{l|cccc}
\hline 
Methods  & PSNR-$\mu$ & PSNR-L & SSIM-$\mu$  & SSIM-L\\
\hline
% MSFFNet-1-scale & 43.9869 & 41.4379 & 0.9906 & 0.9863 \\ 
MSFFNet w/o CA &44.0377 & 41.8150 & 0.9909 & 0.9875 \\ 
MSFFNet w/o MS & 44.1236 & 41.8216 & 0.9908 & 0.9869 \\ % s1
MSFFNet w/o FF & 43.9466 & 41.3520 & 0.9909 & 0.9867 \\ % cos2 161
MSFFNet w/o MSFF & 43.6752 & 41.4698 & 0.9908 & 0.9868  \\ % cos2 136
\hline
MSFFNet w/o FF w/ SPyNet  & 43.9717 &  41.3563 & 0.9905 & 0.9852 \\
MSFFNet w/o FF w/ fixed-pre-trained SPyNet  & 43.6611 &  41.6913 & 0.9896 & 0.9821 \\
MSFFNet w/o FF w/ PWC-Net  & 44.1436 &  42.0084 & 0.9911 & 0.9879 \\ % 161
MSFFNet w/o FF w/ fixed-pre-trained PWC-Net  & 43.3769 &  41.5546 & 0.9891 & 0.9808 \\

MSFFNet  & {44.3298}& 41.8936& 0.9911 &  0.9878\\
\hline

\end{tabular}
\end{center}
\vspace*{-6mm}
\end{table*}
%--------------------------------------
\begin{figure*}[htbp]
\begin{center}
\includegraphics[width=1.0\linewidth]{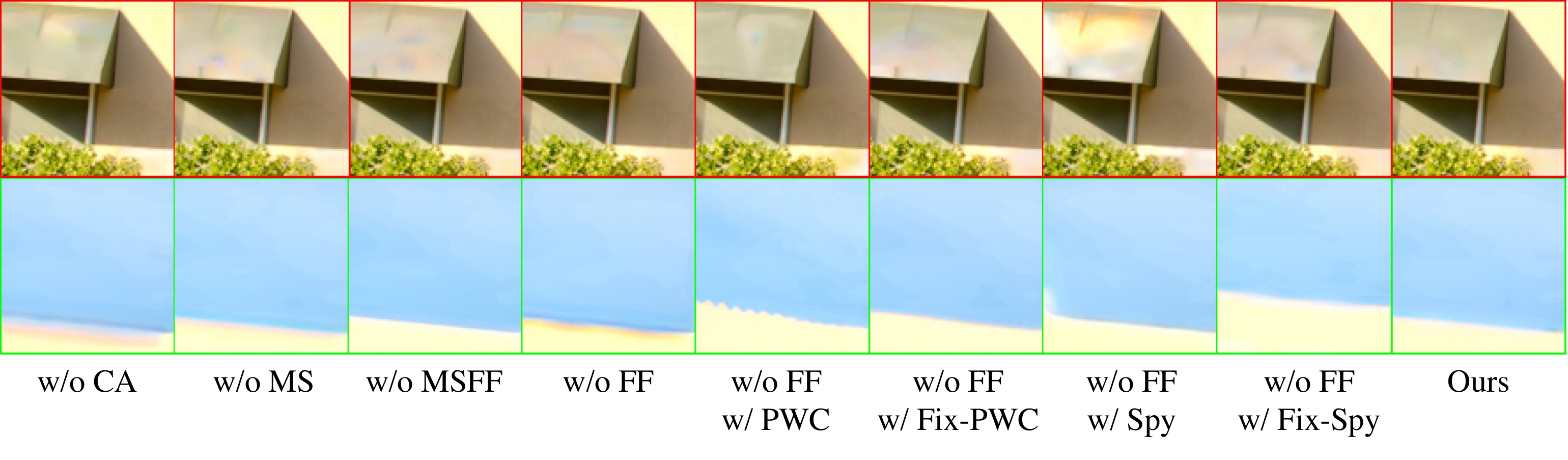}
\end{center}
\vspace*{-6mm}
\caption{Results obtained by ablated networks.}
\label{fig:ab2}
\vspace*{-6mm}
\end{figure*}
% --------------------------------------

%------------------------------------------------------------------------- 
\subsection{Ablation Study}
We demonstrate the effectiveness of each component in the proposed method. We use the same configurations as those used above unless otherwise noted.
\subsubsection{Effects of Optical Flow Learning}
\begin{figure}[htbp]
\begin{center}
\includegraphics[width=.95\linewidth]{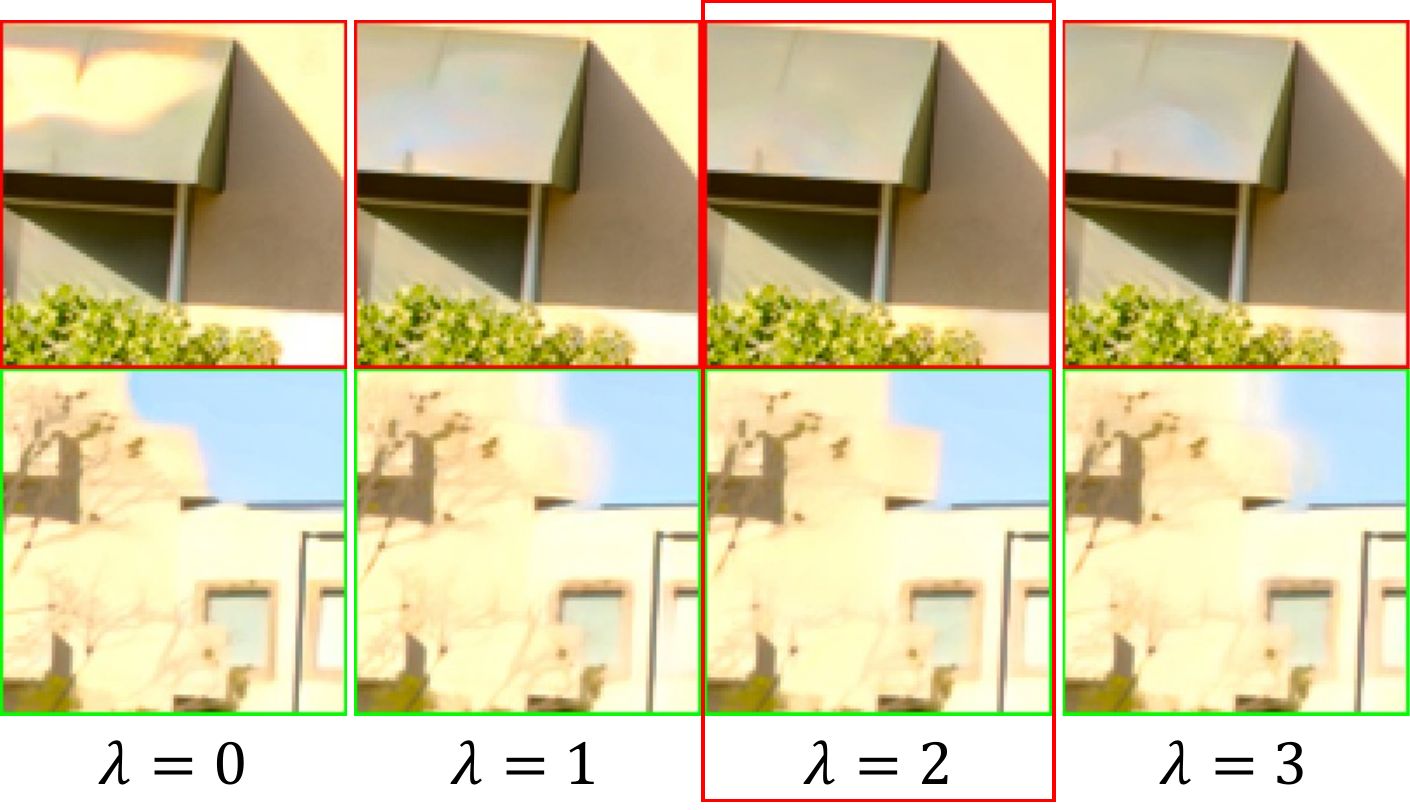}
\end{center}
\vspace*{-4mm}
\caption{Results obtained with different $\lambda$ values in Eq.~\ref{loss-total}. It can be seen that our proposed training (\ie $\lambda > 0$) significantly improve the reconstruction results.}
\label{fig:ab1}
\vspace*{-4mm}
\end{figure}

First, we verify the effect of the proposed loss in Eq.~\ref{loss-reg} by changing the value of the parameter $\lambda$. When $\lambda=0$, the network will be trained only using the $\ell_1$ loss (\ie Eq.~\ref{loss-tm}) between the final outputs and their ground truths. This is equivalent to the previous methodology using optical flow \cite{kalantari2019deep}. It should be noted that although the work \cite{kalantari2019deep} tackles the HDR video reconstruction, the approach also works on the HDR imaging task.
%and does not use the regularized loss in Eq. \ref{loss-reg} to guide the optical flow learning, which means that this ablated network becomes Task-Oriented based model. Since the original Task-Oriented based network \cite{kalantari2019deep} is proposed for HDR video reconstruction which has different settings with HDR imaging, we compare our network with Task-Oriented based network by setting $\lambda=0$.
As shown in Fig.~\ref{fig:ab1}, the model without the proposed loss (\ie $\lambda=0$) generates sever color and geometry distortions. In contrast, the models with the proposed training (\ie $\lambda > 0$) significantly improve the reconstruction results. We also quantitatively evaluate them on the Kalantari \etal’s test set \cite{kalantari2017deep}. As shown in Table~\ref{Table:para}, our proposed training with $\lambda=2$ achieves the best performance.
Even though the gain by the proposed training is not so large, the artifacts usually appear in a small area of an image, and they have only small impacts on these metrics.
%We conclude that a proper $\lambda$ setting can achieve the better performance than other settings including the Task-Oriented based model and too small or large settings do not lead to a good result. 

\begin{table}[htbp]
\begin{center}
  \vspace*{-4mm}
\caption{Results obtained with different $\lambda$ values in Eq.~\ref{loss-total}.}
\label{Table:para} \small
\setlength{\tabcolsep}{5pt}
\begin{tabular}{l|cccc}
\hline 
  & PSNR-$\mu$ & PSNR-L & SSIM-$\mu$  & SSIM-L\\
\hline
$\lambda$=0 & 44.1229 & 41.8947 & 0.9911 & 0.9875 \\ %196
$\lambda$=1 & 44.0978 & 41.7622 & 0.9910 & 0.9868 \\ % 201 to3_cab
% $\lambda$=2.5 & 43.9120 & 41.4850 & 0.9906 & 0.9857  \\ %211
$\lambda$=2  & {44.3298}& 41.8936& 0.9911 &  0.9878 \\ 
$\lambda$=3  & 44.1238 & 41.7043& 0.9911 &  0.9871 \\ %181
\hline
\end{tabular}
\end{center}
\vspace*{-6mm}

\end{table}

%%%%%%%%%%%%%%%%%%%%%%%%%%%%%%%%%%%%%%%%%%%%%%%%%%%%%%%%%%%%%%%%%%%%%%%%%%%%%%%%%%%
%--------------------------------------

%--------------------------------------
\subsubsection{Effects of Different Configurations}
Since our method has several design choices, we conduct experiments to examine which configuration shows the best. Specifically, we examine the effect of channel attention (CA), multi-scale feature flow module (MS-Flow), feature flow module (FF), and multi-scale feature (MS-Fuse). The results are shown in Table~\ref{Table:ablation}. When we eliminate the MS-Flow, we use a single convolution layer to extract feature maps and then concatenate them as $Z_0$. We also do not reconstruct HDR images using the warped feature maps.
When we eliminate the FF, we also do not reconstruct HDR images and directly use $F_{1}^{s}$ as $F_{1\rightarrow r}^{s}$ $(s=1,2,3)$ since there is no optical flow available for the alignment.
%When we eliminate the FF, we also do not reconstruct HDR images and directly use $F_{0}^{s}$ $(s=1,2,3)$ as $F_{1\rightarrow r}^{s}$ $(s=1,2,3)$ since there is no flow used for alignment.
We can observe from Table~\ref{Table:ablation} that CA and MS-Flow are essential to achieve better performance. Figure~\ref{fig:ab2} shows some examples of the zoomed-in patches produced by the models with different configurations. It can be seen that artifacts appear except the MSFFNet (with full components).

%------------------------------------------------------------------------- 
We compare the proposed multi-scale feature flow estimation module with other optical flow methods (SPyNet~\cite{ranjan2017optical} and PWC-Net~\cite{sun2018pwc}). We replace the FF with the optical flow (SPyNet and PWC-Net) in the proposed network named MSFFNet w/o FF w/ SPyNet and MSFFNet w/o FF w/ PWC-Net. As shown in Table~\ref{Table:ablation}, the proposed method achieves better performance than other optical flow-based methods even the PWC-Net has a more complicated network structure (e.g. with correlation layer~\cite{dosovitskiy2015flownet}) than ours. Since our feature flow network can not be pre-trained on other datasets with optical flow ground truth, we also compare the pre-trained optical flow network named MSFFNet w/o FF w/ fixed-pre-trained SPyNet and MSFFNet w/o FF w/ fixed-pre-trained PWC-Net with the model trained on the proposed regularized loss to demonstrate the effectiveness of the proposed regularized loss. These two ablated networks are trained by only using tone-mapped loss in Eq. \ref{loss-tm}. As shown in Table~\ref{Table:ablation}, the model trained on the proposed regularized loss (MSFFNet w/o FF w/ SPyNet and MSFFNet w/o FF w/ PWC-Net) achieve better performance than the pre-trained model (MSFFNet w/o FF w/ fixed-pre-trained SPyNet and MSFFNet w/o FF w/ fixed-pre-trained PWC-Net). As shown in Figure \ref{fig:ab2}, there is severe color and geometry distortion in the images generated from the optical flow-based method.

\section{Conclusions}

In this paper, we propose a new method for generating an HDR image of a dynamic scene from its LDR images along the alignment-before-merging direction. The first step of feature alignment plays a central role in generating high-quality HDR images. Trained by the regularized loss, the multi-scale feature flow module can effectively learn the flow for alignment even in occlusion regions and the large saturated areas, which greatly reduce the artifacts in these regions. After the alignment by the estimated flow, the features from the non-reference image will be fused with the features from the reference image to reconstruct an HDR image. The experimental results have validated the effectiveness of the proposed approach. 

%References are listed in alphabetic order by the surname of the first author, or the identifying word (e.g., in case of a website). Have
%all anonymized references at the beginning of the list.

%%%%%%%%% REFERENCES
{\small
\bibliographystyle{ieee_fullname}
\bibliography{egbib}

\begin{thebibliography}{10}\itemsep=-1pt

\bibitem{anderson2007critters}
Eike~Falk Anderson and Leigh McLoughlin.
\newblock Critters in the classroom: a 3d computer-game-like tool for teaching
  programming to computer animation students.
\newblock In {\em ACM SIGGRAPH 2007 Educators Program}, page~7. 2007.

\bibitem{bogoni2000extending}
Luca Bogoni.
\newblock Extending dynamic range of monochrome and color images through
  fusion.
\newblock In {\em Proceedings 15th International Conference on Pattern
  Recognition. ICPR-2000}, volume~3, pages 7--12. IEEE, 2000.

\bibitem{butler2012naturalistic}
Daniel~J Butler, Jonas Wulff, Garrett~B Stanley, and Michael~J Black.
\newblock A naturalistic open source movie for optical flow evaluation.
\newblock In {\em European conference on computer vision}, pages 611--625.
  Springer, 2012.

\bibitem{chan2021basicvsr}
Kelvin~CK Chan, Xintao Wang, Ke Yu, Chao Dong, and Chen~Change Loy.
\newblock Basicvsr: The search for essential components in video
  super-resolution and beyond.
\newblock In {\em Proceedings of the IEEE/CVF Conference on Computer Vision and
  Pattern Recognition}, pages 4947--4956, 2021.

\bibitem{choi2020pyramid}
Sungil Choi, Jaehoon Cho, Wonil Song, Jihwan Choe, Jisung Yoo, and Kwanghoon
  Sohn.
\newblock Pyramid inter-attention for high dynamic range imaging.
\newblock {\em Sensors}, 20(18):5102, 2020.

\bibitem{dosovitskiy2015flownet}
Alexey Dosovitskiy, Philipp Fischer, Eddy Ilg, Philip Hausser, Caner Hazirbas,
  Vladimir Golkov, Patrick Van Der~Smagt, Daniel Cremers, and Thomas Brox.
\newblock Flownet: Learning optical flow with convolutional networks.
\newblock In {\em Proceedings of the IEEE international conference on computer
  vision}, pages 2758--2766, 2015.

\bibitem{eilertsen2017hdr}
Gabriel Eilertsen, Joel Kronander, Gyorgy Denes, Rafa{\l}~K Mantiuk, and Jonas
  Unger.
\newblock Hdr image reconstruction from a single exposure using deep cnns.
\newblock {\em ACM Transactions on Graphics (TOG)}, 36(6):1--15, 2017.

\bibitem{endoSA2017}
Yuki Endo, Yoshihiro Kanamori, and Jun Mitani.
\newblock Deep reverse tone mapping.
\newblock {\em ACM Transactions on Graphics}, 36(6), 2017.

\bibitem{geiger2012we}
Andreas Geiger, Philip Lenz, and Raquel Urtasun.
\newblock Are we ready for autonomous driving? the kitti vision benchmark
  suite.
\newblock In {\em 2012 IEEE conference on computer vision and pattern
  recognition}, pages 3354--3361. IEEE, 2012.

\bibitem{hafner2014simultaneous}
David Hafner, Oliver Demetz, and Joachim Weickert.
\newblock Simultaneous hdr and optic flow computation.
\newblock In {\em 2014 22nd International Conference on Pattern Recognition},
  pages 2065--2070. IEEE, 2014.

\bibitem{heo2010ghost}
Yong~Seok Heo, Kyoung~Mu Lee, Sang~Uk Lee, Youngsu Moon, and Joonhyuk Cha.
\newblock Ghost-free high dynamic range imaging.
\newblock In {\em Asian Conference on Computer Vision}, pages 486--500.
  Springer, 2010.

\bibitem{hu2013hdr}
Jun Hu, Orazio Gallo, Kari Pulli, and Xiaobai Sun.
\newblock Hdr deghosting: How to deal with saturation?
\newblock In {\em Proceedings of the IEEE Conference on Computer Vision and
  Pattern Recognition}, pages 1163--1170, 2013.

\bibitem{jacobs2008automatic}
Katrien Jacobs, Celine Loscos, and Greg Ward.
\newblock Automatic high-dynamic range image generation for dynamic scenes.
\newblock {\em IEEE Computer Graphics and Applications}, 28(2):84--93, 2008.

\bibitem{kalantari2017deep}
Nima~Khademi Kalantari and Ravi Ramamoorthi.
\newblock Deep high dynamic range imaging of dynamic scenes.
\newblock {\em ACM Transactions on Graphics (TOG)}, 36(4):144:1--144:12, 2017.

\bibitem{kalantari2019deep}
Nima~Khademi Kalantari and Ravi Ramamoorthi.
\newblock Deep hdr video from sequences with alternating exposures.
\newblock In {\em Computer graphics forum}, volume~38, pages 193--205. Wiley
  Online Library, 2019.

\bibitem{kang2003high}
Sing~Bing Kang, Matthew Uyttendaele, Simon Winder, and Richard Szeliski.
\newblock High dynamic range video.
\newblock {\em ACM Transactions on Graphics (TOG)}, 22(3):319--325, 2003.

\bibitem{khan2006ghost}
Erum~Arif Khan, Ahmet~Oguz Akyuz, and Erik Reinhard.
\newblock Ghost removal in high dynamic range images.
\newblock In {\em 2006 International Conference on Image Processing}, pages
  2005--2008. IEEE, 2006.

\bibitem{KingmaB14}
Diederik~P Kingma and Jimmy Ba.
\newblock Adam: {A} method for stochastic optimization.
\newblock In {\em 3rd International Conference on Learning Representations,
  {ICLR} 2015}, 2015.

\bibitem{lee2014ghost}
Chul Lee, Yuelong Li, and Vishal Monga.
\newblock Ghost-free high dynamic range imaging via rank minimization.
\newblock {\em IEEE Signal Processing Letters}, 21(9):1045--1049, 2014.

\bibitem{lin2021fdan}
Jiayi Lin, Yan Huang, and Liang Wang.
\newblock Fdan: Flow-guided deformable alignment network for video
  super-resolution.
\newblock {\em arXiv preprint arXiv:2105.05640}, 2021.

\bibitem{loshchilov2016sgdr}
Ilya Loshchilov and Frank Hutter.
\newblock {SGDR:} stochastic gradient descent with warm restarts.
\newblock In {\em 5th International Conference on Learning Representations,
  {ICLR} 2017}, 2017.

\bibitem{mantiuk2011hdr}
Rafa{\l} Mantiuk, Kil~Joong Kim, Allan~G Rempel, and Wolfgang Heidrich.
\newblock Hdr-vdp-2: A calibrated visual metric for visibility and quality
  predictions in all luminance conditions.
\newblock {\em ACM Transactions on Graphics (TOG)}, 30(4):1--14, 2011.

\bibitem{paszke2017automatic}
Adam Paszke, Sam Gross, Soumith Chintala, Gregory Chanan, Edward Yang, Zachary
  DeVito, Zeming Lin, Alban Desmaison, Luca Antiga, and Adam Lerer.
\newblock Automatic differentiation in pytorch.
\newblock 2017.

\bibitem{poynton2012digital}
Charles Poynton.
\newblock {\em Digital video and HD: Algorithms and Interfaces}.
\newblock Elsevier, 2012.

\bibitem{prabhakar2020towards}
K~Ram Prabhakar, Susmit Agrawal, Durgesh~Kumar Singh, Balraj Ashwath, and
  R~Venkatesh Babu.
\newblock Towards practical and efficient high-resolution hdr deghosting with
  cnn.
\newblock In {\em European Conference on Computer Vision}, pages 497--513.
  Springer, 2020.

\bibitem{prabhakar2019fast}
K~Ram Prabhakar, Rajat Arora, Adhitya Swaminathan, Kunal~Pratap Singh, and
  R~Venkatesh Babu.
\newblock A fast, scalable, and reliable deghosting method for extreme exposure
  fusion.
\newblock In {\em 2019 IEEE International Conference on Computational
  Photography (ICCP)}, pages 1--8. IEEE, 2019.

\bibitem{pu2020robust}
Zhiyuan Pu, Peiyao Guo, M~Salman Asif, and Zhan Ma.
\newblock Robust high dynamic range (hdr) imaging with complex motion and
  parallax.
\newblock In {\em Proceedings of the Asian Conference on Computer Vision},
  2020.

\bibitem{ranjan2017optical}
Anurag Ranjan and Michael~J Black.
\newblock Optical flow estimation using a spatial pyramid network.
\newblock In {\em Proceedings of the IEEE conference on computer vision and
  pattern recognition}, pages 4161--4170, 2017.

\bibitem{rempel2009video}
Allan~G Rempel, Wolfgang Heidrich, Hiroe Li, and Rafa{\l} Mantiuk.
\newblock Video viewing preferences for hdr displays under varying ambient
  illumination.
\newblock In {\em Proceedings of the 6th Symposium on Applied Perception in
  Graphics and Visualization}, pages 45--52, 2009.

\bibitem{scharstein2014high}
Daniel Scharstein, Heiko Hirschm{\"u}ller, York Kitajima, Greg Krathwohl, Nera
  Ne{\v{s}}i{\'c}, Xi Wang, and Porter Westling.
\newblock High-resolution stereo datasets with subpixel-accurate ground truth.
\newblock In {\em German conference on pattern recognition}, pages 31--42.
  Springer, 2014.

\bibitem{sen2012robust}
Pradeep Sen, Nima~Khademi Kalantari, Maziar Yaesoubi, Soheil Darabi, Dan~B
  Goldman, and Eli Shechtman.
\newblock Robust patch-based hdr reconstruction of dynamic scenes.
\newblock {\em ACM Transactions on Graphics (TOG)}, 31(6):203:1--203:12, 2012.

\bibitem{sun2018pwc}
Deqing Sun, Xiaodong Yang, Ming-Yu Liu, and Jan Kautz.
\newblock Pwc-net: Cnns for optical flow using pyramid, warping, and cost
  volume.
\newblock In {\em Proceedings of the IEEE conference on computer vision and
  pattern recognition}, pages 8934--8943, 2018.

\bibitem{tursun2016objective}
Okan~Tarhan Tursun, Ahmet~O{\u{g}}uz Aky{\"u}z, Aykut Erdem, and Erkut Erdem.
\newblock An objective deghosting quality metric for hdr images.
\newblock In {\em Computer Graphics Forum}, volume~35, pages 139--152. Wiley
  Online Library, 2016.

\bibitem{wang2018non}
Xiaolong Wang, Ross Girshick, Abhinav Gupta, and Kaiming He.
\newblock Non-local neural networks.
\newblock In {\em Proceedings of the IEEE conference on computer vision and
  pattern recognition}, pages 7794--7803, 2018.

\bibitem{wu2018deep}
Shangzhe Wu, Jiarui Xu, Yu-Wing Tai, and Chi-Keung Tang.
\newblock Deep high dynamic range imaging with large foreground motions.
\newblock In {\em Proceedings of the European Conference on Computer Vision},
  pages 117--132, 2018.

\bibitem{xue2019video}
Tianfan Xue, Baian Chen, Jiajun Wu, Donglai Wei, and William~T Freeman.
\newblock Video enhancement with task-oriented flow.
\newblock {\em International Journal of Computer Vision}, 127(8):1106--1125,
  2019.

\bibitem{yan2019attention}
Qingsen Yan, Dong Gong, Qinfeng Shi, Anton van~den Hengel, Chunhua Shen, Ian
  Reid, and Yanning Zhang.
\newblock Attention-guided network for ghost-free high dynamic range imaging.
\newblock In {\em Proceedings of the IEEE Conference on Computer Vision and
  Pattern Recognition}, pages 1751--1760, 2019.

\bibitem{yan2017high}
Qingsen Yan, Jinqiu Sun, Haisen Li, Yu Zhu, and Yanning Zhang.
\newblock High dynamic range imaging by sparse representation.
\newblock {\em Neurocomputing}, 269:160--169, 2017.

\bibitem{yan2020deep}
Qingsen Yan, Lei Zhang, Yu Liu, Yu Zhu, Jinqiu Sun, Qinfeng Shi, and Yanning
  Zhang.
\newblock Deep hdr imaging via a non-local network.
\newblock {\em IEEE Transactions on Image Processing}, 29:4308--4322, 2020.

\bibitem{ye2021progressive}
Qian Ye, Jun Xiao, Kin-man Lam, and Takayuki Okatani.
\newblock Progressive and selective fusion network for high dynamic range
  imaging.
\newblock In {\em Proceedings of the ACM International Conference on
  Multimedia}, 2021.

\bibitem{yu2016back}
Jason~J Yu, Adam~W Harley, and Konstantinos~G Derpanis.
\newblock Back to basics: Unsupervised learning of optical flow via brightness
  constancy and motion smoothness.
\newblock In {\em European Conference on Computer Vision}, pages 3--10.
  Springer, 2016.

\bibitem{zhang2011gradient}
Wei Zhang and Wai-Kuen Cham.
\newblock Gradient-directed multiexposure composition.
\newblock {\em IEEE Transactions on Image Processing}, 21(4):2318--2323, 2011.

\bibitem{zhang2018image}
Yulun Zhang, Kunpeng Li, Kai Li, Lichen Wang, Bineng Zhong, and Yun Fu.
\newblock Image super-resolution using very deep residual channel attention
  networks.
\newblock In {\em Proceedings of the European conference on computer vision
  (ECCV)}, pages 286--301, 2018.

\bibitem{zhao2016loss}
Hang Zhao, Orazio Gallo, Iuri Frosio, and Jan Kautz.
\newblock Loss functions for image restoration with neural networks.
\newblock {\em IEEE Transactions on Computational Imaging}, 3(1):47--57, 2016.

\bibitem{zhu2019deformable}
Xizhou Zhu, Han Hu, Stephen Lin, and Jifeng Dai.
\newblock Deformable convnets v2: More deformable, better results.
\newblock In {\em Proceedings of the IEEE/CVF Conference on Computer Vision and
  Pattern Recognition}, pages 9308--9316, 2019.

\end{thebibliography}
}

\end{document}